\documentclass[conference,10pt,letterpaper]{IEEEtran}
\usepackage{fontspec}
\usepackage{amsmath,amssymb,bm}
\usepackage{graphicx,booktabs,array,tabularx,makecell}
\usepackage{cite,placeins}
\usepackage[hidelinks,unicode]{hyperref}
\usepackage{url}
\newcolumntype{Y}{>{\raggedright\arraybackslash}X}
\title{High-Precision Docking of Differential-Drive Mobile Robots Through LiDAR and Vision Collaboration}
\author{\IEEEauthorblockN{Zaide Zhang}\IEEEauthorblockA{NovaXBot\\\href{mailto:zhangzaide7@gmail.com}{zhangzaide7@gmail.com}}\and\IEEEauthorblockN{Zheyuan Zhang}\IEEEauthorblockA{Beijing Institute of Technology\\\href{mailto:1120230155@bit.edu.cn}{1120230155@bit.edu.cn}}}
\date{}
\hypersetup{pdftitle={High-Precision Docking of Differential-Drive Mobile Robots Through LiDAR and Vision Collaboration},pdfauthor={Zaide Zhang and Zheyuan Zhang}}
\begin{document}
\maketitle

\begin{abstract}
Precise docking requires a differential-drive robot to correct lateral offset through coupled rotation and translation within a short approach distance. We combine global navigation with terminal control using complementary LiDAR and visual observations. A line fitted to the cabinet front gives normal distance and heading. An AprilTag identifies the workstation; intersecting its center ray with the LiDAR plane determines tangential position. Taught observations define the docking reference and can cancel repeatable fixed biases under unchanged mounting and geometry. Lateral pre-alignment and continuous visual servoing coordinate the approach, with predictive stopping to account for low-speed actuation lag. Across two sessions using the same navigation and docking code at five workstations on Robot 2, all 111 tasks completed, including 101 on the first attempt. Relative to the taught observations, onboard longitudinal, lateral and heading residual root-mean-square values were 1.51\,mm, 5.93\,mm and $0.238^\circ$, respectively. Joint acceptance within $\pm10$\,mm per position axis and $\pm1^\circ$ heading was 96.4\% (107/111), or 95.5\% (106/111) with a $\pm0.5^\circ$ heading limit.
\end{abstract}

\begin{IEEEkeywords}
AprilTag, differential drive, LiDAR--vision collaboration, mobile robot docking, teaching, visual servoing
\end{IEEEkeywords}

\section{Introduction}
A differential-drive robot cannot translate sideways. Millimeter-scale lateral alignment at a workstation must therefore be achieved through coupled rotation and translation, while the remaining distance for correction shrinks near the endpoint. Turning in place can introduce additional lateral slip. For food-service robots retrieving ingredients or operating equipment at cabinet fronts, the controller must align position and heading within a confined approach region. Reaching a nearby map goal does not by itself establish the required pose relative to the physical interface.

Simultaneous localization and mapping (SLAM)\cite{r1}, SLAM Toolbox\cite{r2}, and Navigation2\cite{r3} support travel between workstations, with local motion controlled through methods such as model predictive control\cite{r4} and regulated pure pursuit (RPP)\cite{r5}. Terminal docking also requires observations of the workstation. V-shaped features\cite{r6} and reflective markers\cite{r7} make LiDAR targets identifiable but require dedicated geometry or reflective material. Fitting a scan line to an ordinary cabinet front yields normal distance and heading. Translation along that surface leaves the line unchanged, so tangential position requires another reference.

AprilTag\cite{r8,r9} identifies the workstation and establishes a visual position reference. Perspective-$n$-point (PnP) methods, including infinitesimal plane-based pose estimation (IPPE)\cite{r10}, estimate pose from planar image features; refined tag edges improve feature measurement\cite{r11}. As the projected tag shrinks with range, pixel perturbations propagate differently into lateral position and depth. We retain the tag-center bearing and intersect its ray with the LiDAR surface, assigning heading feedback to LiDAR.

We implement this allocation with a taught reference and a terminal controller on Robot 2. Across two sessions with the same navigation and docking code, all 111 tasks completed and 107 met a unified onboard feedback residual criterion of $\pm10$\,mm per position axis and $\pm1^\circ$ in heading. The contributions are:
\begin{enumerate}
\item A local three-degree-of-freedom feedback construction that combines surface normal distance and heading with tag-based tangential position, including a taught zero-residual condition and distance-dependent compensation.
\item A differential-drive docking controller that combines affine lateral pre-alignment, distance-scheduled steering, and predictive stopping.
\item A workstation-level evaluation of 111 same-version tasks that distinguishes completion, first-attempt success, and residual acceptance.
\end{enumerate}

\section{Related Work}
\subsection{LiDAR Targets and Local Geometry}
V-shaped features\cite{r6}, intensity markers\cite{r7}, and passive guides\cite{r19} make relative pose observable through identifiable workstation structure. Multistage localization also couples global positioning with scan matching in the docking region\cite{r15}. We fit an existing planar cabinet front using random sample consensus (RANSAC)\cite{r18} and orthogonal refinement to estimate normal distance and heading. The tag bearing resolves the remaining translation along the surface, avoiding the need for a dedicated workstation profile.

\subsection{Visual Fiducials and Multisensor Docking}
AprilTag detection and layout methods\cite{r8,r9,r20} and square fiducials under occlusion\cite{r25} enable workstation identification; planar pose estimation\cite{r10,r11} recovers pose from image features. Dual tags have also supported LiDAR localization\cite{r21}, while camera calibration determines the pixel-to-ray mapping\cite{r24}. Docking systems have combined visual recognition with LiDAR\cite{r12,r13}, or depth-aided tag localization with LiDAR heading correction\cite{r14}. We instead compute tangential position from the intersection of the tag-center ray and the LiDAR plane. PnP initializes the search region and gates range and reprojection quality; heading feedback comes from the fitted LiDAR line.

\subsection{Taught References and Differential-Drive Control}
Visual teach and repeat stores observations along a route\cite{r22}; visual servoing updates motion from desired image features or pose errors\cite{r16,r23}; smooth pose control addresses differential-drive motion constraints\cite{r17}. Dedicated calibration can correct systematic wheel and wheelbase errors\cite{r27}. We instead define a fixed workstation endpoint through taught observations, compensate range-dependent tangential drift during approach, and learn the gain and offset of lateral correction maneuvers from local measurements. The resulting controller targets endpoint reproduction and approach alignment. Table~\ref{tab:comparison} summarizes the sensing requirements and reported measurements of representative docking systems.

\begin{table*}[!t]
\caption{Docking approaches and measurement scope. Metrics and test conditions differ across studies; values do not constitute a common-benchmark ranking.}
\label{tab:comparison}\centering\footnotesize\renewcommand{\arraystretch}{1.18}
\begin{tabularx}{\textwidth}{YYYY}\toprule
Method & Sensors & Extra hardware / target & Reported result and measurement \\ \midrule
V-shaped features\cite{r6} & LiDAR & V-shaped workstation profile & Approx. $\pm20$\,mm ($x/y$), $\pm1^\circ$; two-stage physical tests; independent instrument and RMSE not specified \\
Intensity markers\cite{r7} & LiDAR, odometry, ultrasound & Reflective markers & 200 trials per scenario; docking success 99.5\% / 98.5\%; charging-bar positions inspected; no terminal pose-error statistics \\
Jia et al.\cite{r13} & Camera, LiDAR & Visually recognized charging station & No terminal error reported; 95\% is charging-station image recognition accuracy; Gazebo and Husky demonstrations \\
Dai and Lee\cite{r14} & Camera, depth, LiDAR & AprilTag & No terminal error statistics; ruler used for pixel-to-3-D validation; docking convergence shown in Gazebo \\
This work & IR camera, 2-D LiDAR & Ordinary cabinet front and fixed AprilTag & Onboard planar residual root-mean-square error (RMSE) 6.12\,mm; 111 taught-observation reproduction tasks \\
\bottomrule\end{tabularx}\end{table*}

\section{Method}
\label{sec:method}
\begin{figure*}[!t]
\centering
\includegraphics[width=.83\textwidth,height=.70\textheight,keepaspectratio]{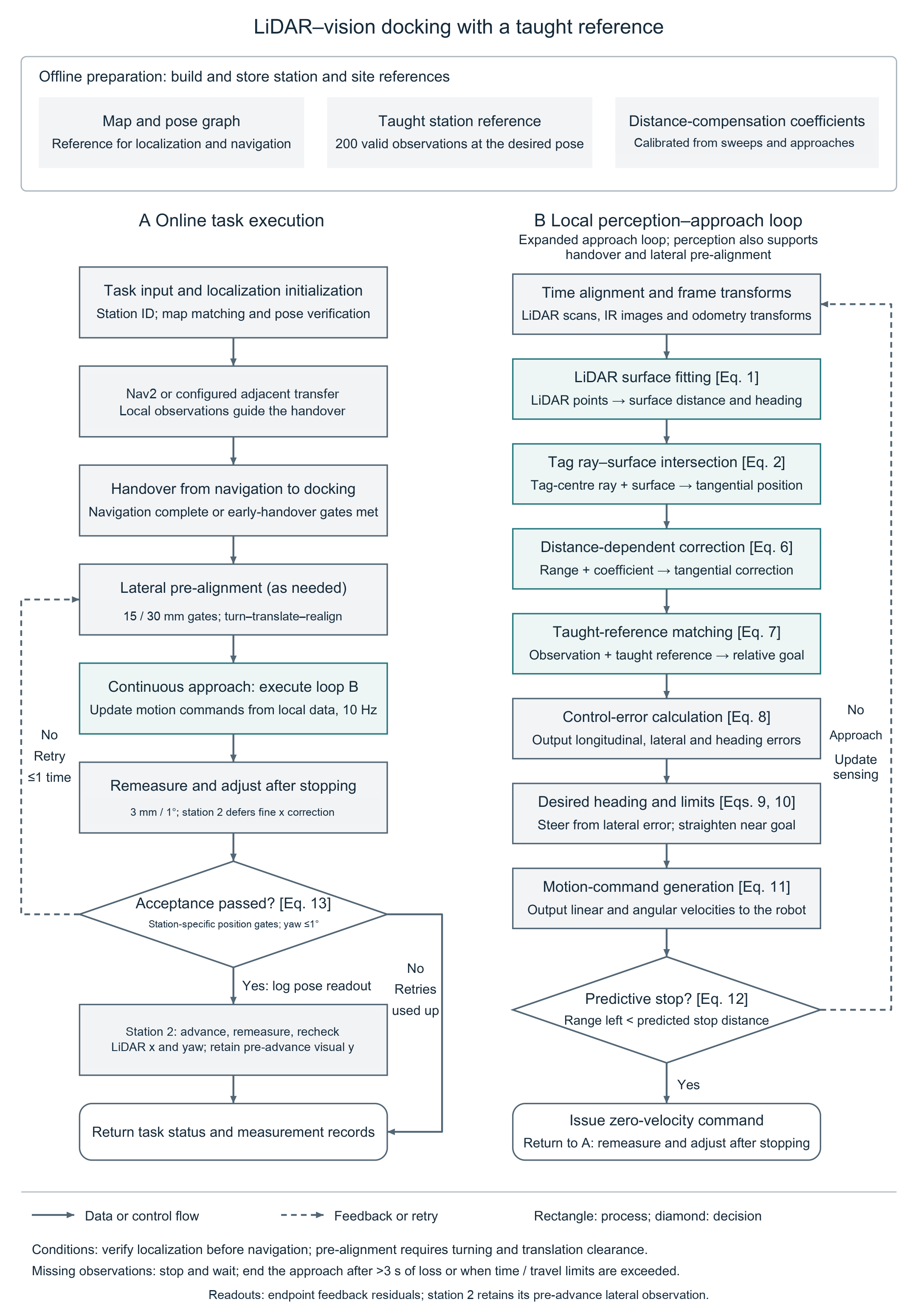}
\caption{LiDAR–vision docking with a taught reference. Finishing settings shown here correspond to the implementation revised on 3 November; Section~\ref{sec:versions} specifies the historical settings used in the primary experiments.}
\label{fig:1}
\end{figure*}

Each workstation has a visible vertical cabinet front, a uniquely identified AprilTag and a taught reference. Global navigation brings the robot into the approach region. Local control starts only after the target observation is verified and navigation has stopped, then proceeds through lateral pre-alignment and continuous approach (Fig.~\ref{fig:1}). We write ${}^{A}\bm T_B$ for the transform that maps coordinates from frame $B$ into frame $A$; Table~\ref{tab:1} defines the frames. Implementation settings are listed in Appendix~\ref{app:params}, with the experimental version identified in Section~\ref{sec:versions}.
\begin{table}[!t]
\caption{Coordinate frames and principal notation.}
\label{tab:1}
\centering
\footnotesize
\setlength{\tabcolsep}{3pt}
\renewcommand{\arraystretch}{1.18}
\begin{tabularx}{\columnwidth}{lY}
\toprule
Symbol & Definition \\
\midrule
M & Global map frame \\
O & Odometry frame \\
B & Robot body frame base\_link \\
C & Infrared camera frame \\
F & Workstation surface frame \\
G & Taught goal frame \\
$\bm g=[g_x,g_y]^\top,\varphi$ & Goal position and heading in B \\
$\bm e=[e_x,e_y]^\top,\psi$ & Robot position and heading in G \\
\bottomrule
\end{tabularx}
\end{table}
\subsection{DOF Allocation and Ray--Plane Intersection}
\label{sec:geometry}
A fitted cabinet line determines normal distance and heading. Tangential position comes from the tag center. To align the observations, we transform the image and LiDAR observations into odometry frame $O$ at their respective acquisition times, applying a fixed image-time offset. RANSAC selects the dominant surface line near the initial tag position; orthogonal least squares then refines the fit. For inliers $\bm p_i$, centroid $\bm c$, and unit normal $\bm n$ pointing toward the LiDAR, the objective is
\begin{equation}
\min_{\bm c,\bm n}\sum_{i\in I}\left[\bm n^\top(\bm p_i-\bm c)\right]^2,\quad \lVert\bm n\rVert=1
\label{eq:2}
\end{equation}
After normalizing infrared intensity, we detect the tag and refine its corners. The quadrilateral diagonals intersect at the tag center, whose undistorted homogeneous coordinates are $\tilde{\bm u}=[u,v,1]^\top$. Let $\bm K$ be the camera intrinsic matrix and $({}^{O}\bm R_C,{}^{O}\bm t_C)$ its pose in $O$. We extend the fitted LiDAR line vertically to form a plane. The ray origin, direction, and intersection are
\begin{equation}
\begin{aligned}\bm o&={}^{O}\bm t_C,\quad \bm r={}^{O}\bm R_C\bm K^{-1}\tilde{\bm u},\\\lambda&=-\frac{\bm n^\top(\bm o_{xy}-\bm c)}{\bm n^\top\bm r_{xy}},\quad\bm q=\bm o+\lambda\bm r.\end{aligned}
\label{eq:3}
\end{equation}
The horizontal intersection sets the origin of surface frame $F$. Its $x$-axis points into the surface and its $y$-axis points leftward along it. Backward or nearly plane-parallel rays are rejected, along with observations that fail identity, reprojection, fit-quality or temporal-continuity checks. We take heading from the LiDAR line and use PnP for initialization and quality gating. Depth is inherited from the fitted plane, not independently measured by the intersection.

\paragraph{Lateral error propagation}
Let $D$ denote the camera-to-plane normal distance and $\beta$ the horizontal bearing relative to the plane normal. The tangential intersection is $s=D\tan\beta$, with first-order perturbation
\begin{equation}
\delta s=\tan\beta\,\delta D+D\sec^2\beta\,\delta\beta.
\label{eq:propagation}
\end{equation}
For a local model with the camera optical axis normal to the surface, $\tan\beta=(u-c_x)/f_x$. With independent pixel, distance, and effective angular errors,
\begin{equation}
\sigma_{s,\mathrm{ray}}^2\simeq
(D/f_x)^2\sigma_u^2+\tan^2\beta\,\sigma_D^2
+D^2\sec^4\beta\,\sigma_\alpha^2.
\label{eq:rayvariance}
\end{equation}
Here, $\sigma_\alpha$ represents uncertainty in extrinsic yaw and plane-normal direction; correlated terms require their covariances. For a frontoparallel tag of side length $L$ and image width $w\simeq f_xL/Z$, a local approximation to the depth sensitivity in PnP is
\begin{equation}
\begin{aligned}
\delta Z&\simeq-\frac{Z^2}{f_xL}\delta w,\\
\delta s_{\mathrm{PnP}}&\simeq\frac{Z}{f_x}\delta u
+\frac{u-c_x}{f_x}\delta Z.
\end{aligned}
\label{eq:pnpvariance}
\end{equation}
Table~\ref{tab:2} gives $L=0.150$\,m and a $640\times400$ image, so the depth sensitivity is $Z^2/(0.150f_x)$. Away from the principal point, the second term couples depth uncertainty into lateral position. Ray--plane intersection replaces this depth estimate with the LiDAR plane constraint. Near the principal point, both methods have the same leading pixel term, so their relative uncertainty depends on the remaining terms. A full IPPE/PnP comparison also requires corner correlations and orientation coupling.

We checked local propagation against early static records from Robot 2 at P5 on 22 October. The recorded intrinsics were $f_x=f_y=196.691$\,px with principal point $(320.529,198.852)$\,px. Fourteen valid stops contained 370 frames (15--30 per stop), with mean cabinet distances of 0.456--0.850\,m. Within-stop sample standard deviations were 0.022--0.103\,px for horizontal image position and 0.246--0.364\,mm for normal distance. Plane-heading variability ranged from $0.048$ to $0.074^\circ$, and tangential variability from 0.704 to 1.218\,mm. For the full ray--plane mapping, let $\bm J$ be the local Jacobian and $\bm\Sigma$ the sample covariance of $\bm z=(u,v,d,\theta)$. Retaining cross-covariances gives $\sigma_s=\sqrt{\bm J\bm\Sigma\bm J^\mathsf{T}}=0.705$--$1.221$\,mm. These static records measure short-term variability at fixed extrinsics, separately from the 111-task cohort of 2 November; systematic extrinsic bias remains outside this check.
\subsection{Taught Zero-Residual Condition and Distance Compensation}
Teaching starts with the robot at an acceptable docking pose. We average valid translations arithmetically and orientations circularly to obtain ${}^{B^*}\bm T_F$, where $B^*$ is the taught body frame. During approach, the tangential reading $s(t)$ varies with surface distance $d(t)$. A workstation-specific slope $a_j$ compensates this drift:
\begin{equation}
\begin{aligned}\Delta s_j(t)&=-a_j\left[d(t)-d_j^*\right],\\s_{\mathrm{corr}}(t)&=s(t)+\Delta s_j(t).\end{aligned}
\label{eq:4}
\end{equation}
Distance sweeps initialize $a_j$; approach trajectories refine it after accounting for lateral motion and heading change. The current P1--P5 coefficients are $+0.058,+0.069,-0.025,-0.028,-0.022$\,m/m. At the taught distance $d_j^*$, the compensation is zero. The goal frame $G$ represents the taught body pose, so $({}^{B^*}\bm T_F)^{-1}$ maps goal-frame coordinates into the surface frame. Applying the tangential correction there and then transforming into $O$ gives
\begin{equation}
\begin{aligned}{}^{O}\bm T_G(t)&={}^{O}\bm T_F(t)\operatorname{Trans}_y(\Delta s_j(t))\left({}^{B^*}\bm T_F\right)^{-1},\\{}^{B}\bm T_G(t)&=\left({}^{O}\bm T_B(t)\right)^{-1}{}^{O}\bm T_G(t).\end{aligned}
\label{eq:5}
\end{equation}
Here, $\operatorname{Trans}_y$ translates along the surface frame. At the taught pose, the current and taught observations coincide and compensation vanishes; the transform chain reduces to ${}^{B}\bm T_G=\bm I$. Repeatable fixed biases can therefore cancel at this reference pose under unchanged mounting and geometry. Away from it, error scaling depends on ray direction and calibration.

Equation~\eqref{eq:propagation} also tests whether fixed yaw miscalibration can explain $a_j$. Near normal incidence, with normal translation and consistent signs, a yaw error $\delta\alpha$ produces a drift slope $a\simeq\tan\delta\alpha$. The five coefficients imply approximately $+3.32^\circ,+3.95^\circ,-1.43^\circ,-1.60^\circ,-1.26^\circ$. One fixed yaw cannot explain both sign groups under this simplified model. Different viewing directions, plane fitting, and residual motion coupling require synchronized observations to separate. We therefore retain empirical workstation-specific coefficients without identifying them as measured camera--LiDAR extrinsic errors.
\subsection{Lateral Allocation and Predictive Stopping}
\label{sec:control}
Before continuous approach, the robot aligns its heading to the LiDAR line. We then evaluate lateral error in the goal frame and use a turn--translate--realign maneuver for larger offsets. To account for the maneuver's response, we fit $\Delta y=ku+c$ to observed error removal $\Delta y$ and commanded lateral shift $u$. A requested correction $q$ yields $u=\operatorname{clip}((q-c)/k;-u_{\max},u_{\max})$. The gain $k$ scales the commanded shift; the offset $c$ accounts for displacement caused by turning. Learning windows and bounds are listed in Appendix~\ref{app:params}.

Continuous servoing converts body-frame goal pose $(\bm g,\varphi)$ into goal-frame robot error $(\bm e,\psi)$, separating remaining forward distance $\ell$ from lateral error:
\begin{equation}
\begin{aligned}\begin{bmatrix}e_x\\e_y\end{bmatrix}&=-\bm R(\varphi)^\top\begin{bmatrix}g_x\\g_y\end{bmatrix},\\\psi&=\operatorname{wrap}(-\varphi),\quad\ell=-e_x.\end{aligned}
\label{eq:6}
\end{equation}
Define $\operatorname{clip}(z;a,b)=\min[b,\max(a,z)]$. Lateral correction consumes forward travel: $\dot e_y=v\sin\psi$. As the remaining distance shrinks, we reduce the allowable steering angle and finally command a straight approach:
\begin{equation}
\psi_d=\begin{cases}\operatorname{clip}(-k_y(e_y-\eta_j);-\psi_{\lim},\psi_{\lim}),&\ell\geq\ell_{\mathrm{straight}},\\0,&\ell<\ell_{\mathrm{straight}}.\end{cases}
\label{eq:7}
\end{equation}
\begin{equation}
\psi_{\lim}(\ell)=\min\left[\psi_{\mathrm{cap}},\max(\psi_{\min},\gamma\ell)\right]
\label{eq:8}
\end{equation}
Only P2 uses the lateral steering target $\eta_j$ to compensate reading drift over the final straight segment. Before that segment, steering corrects lateral offset; afterwards, the desired heading is zero. The gains $k_y$, $k_\psi$ and $k_v$ act on lateral, heading and distance errors. Steering limits are set by $\psi_{\mathrm{cap}}$ and $\omega_{\lim}$, while $\gamma$ scales the heading bound with distance. The transition occurs at $\ell_{\mathrm{straight}}$. Velocity commands are
\begin{equation}
\begin{aligned}\omega&=\operatorname{clip}\left[k_\psi(\psi_d-\psi);-\omega_{\lim},\omega_{\lim}\right],\\v&=\max\left[v_{\min},\min(v_{\max},k_v\ell)\right].\end{aligned}
\label{eq:9}
\end{equation}
The robot continues moving briefly after a zero-velocity command. We estimate this travel from the preceding command $v_{k-1}$ and stop when
\begin{equation}
\ell<\max\left(\ell_{\mathrm{stop}},v_{k-1}\tau_{\mathrm{stop}}\right).
\label{eq:10}
\end{equation}
Once the robot settles, fresh observations guide heading correction followed by longitudinal correction. Observation loss stops motion. If the endpoint fails acceptance and a retry is permitted, the robot retreats to the staging pose and approaches again. Acceptance uses the componentwise means of valid endpoint observations:
\begin{equation}
|\bar g_x|\leq\varepsilon_j,\quad|\bar g_y|\leq\varepsilon_j,\quad|\bar\varphi|\leq1^{\circ}
\label{eq:11}
\end{equation}
The operational gate $\varepsilon_j$ is 10\,mm at P1--P4 and 20\,mm at P5; retrospective evaluation applies 10\,mm per axis everywhere. P2 adds a LiDAR-guided advance after reaching the taught pose. Its final $x$ and heading are remeasured, while $y$ retains the pre-advance visual reading, forming a composite endpoint record. Appendix~\ref{app:params} gives versioned settings and sample requests; Section~\ref{sec:residuals} defines the analysis fields.

\section{Experimental Setup}

\subsection{Experimental Platform and Workstations}

The experiments evaluated LiDAR–vision docking on Robot 2, a two-wheel differential-drive platform operating at five fixed workstations in the site\_b environment. The workstations were arranged along two cabinet rows; P3–P5 shared one row. Each workstation carried a unique tag36h11 AprilTag (IDs 20–24; black-border side length 150\,mm) and a taught docking reference. LiDAR supplied cabinet-normal distance and heading, while the infrared image from the left camera supplied the tag-center observation for along-surface positioning.

Thor executed global navigation, terminal perception and docking supervision; the ECU handled chassis actuation and sensor acquisition. The software used ROS 2 Jazzy, slam\_toolbox localization and Nav2 with regulated pure pursuit. Table~\ref{tab:2} lists the documented operating configuration. The image and odometry rates are observed operating rates rather than accuracy specifications.

\begin{table}[!t]
\caption{Docking platform configuration.}
\label{tab:2}
\centering
\footnotesize
\setlength{\tabcolsep}{3pt}
\renewcommand{\arraystretch}{1.18}
\begin{tabularx}{\columnwidth}{lY}
\toprule
Component & Configuration \\
\midrule
Drive and odometry & Differential drive; wheel–IMU fusion, $\approx 98$\,Hz \\
LiDAR & LDS-E210-E; $\approx 15$\,Hz; 180\ensuremath{^{\circ}}; 0.15\ensuremath{^{\circ}} per beam \\
Cameras & Two Orbbec DaBai Max cameras; left IR 640 \ensuremath{\times} 400, $\approx 10$\,Hz \\
Tags & tag36h11; 150\,mm side; P1–P5 correspond to IDs 20–24 \\
Computing & Thor and ECU; gigabit Ethernet \\
Software & ROS 2 Jazzy; slam\_toolbox; Nav2 RPP \\
\bottomrule
\end{tabularx}
\end{table}

Trials were conducted at fixed indoor workstations with emergency-stop supervision; the Limitations section specifies the safety configuration.

\subsection{Trial Protocol and Version Grouping}
\label{sec:versions}

A trial consisted of a dispatched docking goal, any associated departure, coarse transfer, local alignment and final verification. Start and goal identifiers, task timestamps, terminal readings, attempt counts, return codes and log identifiers were retained. Trials followed operational routes rather than a randomized, balanced design. The primary analysis combined two sessions on 2 November 2026 that used the same navigation and docking code: nine trials after startup and 102 executed goals from the platform-driven pressure test. The latter also contained 29 requests rejected before motion because the requested route was prohibited. These requests were reported separately and excluded from the executed-trial denominator.

The cohort contained 111 unique logged trials: 27, 28, 28, 1 and 27 at P1--P5, respectively. The pressure-test route repeatedly visited P1, P2, P3 and P5; P4 appeared in the startup session. Earlier sessions and the 3 November speed revisions are reported separately in Table~\ref{tab:3}; their error distributions and session totals were not pooled with the primary cohort.

The navigation and docking code remained unchanged during the primary cohort, while learned runtime state continued to update. This state includes staging offsets, lateral-action models, adjacent-transfer corrections and the P2 steering target. The cohort therefore evaluates online operation. On 2 November, the finishing heading target was $0.5^\circ$ and final acceptance used $1^\circ$; the P2 final advance used 0.02\,m/s with an 8\,mm margin. The implementation in Fig.~\ref{fig:1} and Appendix~\ref{app:params} includes the later speed revisions.

\begin{table}[!t]
\caption{Recorded trial groups.}
\label{tab:3}
\centering
\footnotesize
\setlength{\tabcolsep}{3pt}
\renewcommand{\arraystretch}{1.18}
\begin{tabularx}{\columnwidth}{p{.32\columnwidth}YYYY}
\toprule
Group & Records & Rejected & Executed & Completed \\
\midrule
2 Nov startup & 9 & 0 & 9 & 9 \\
2 Nov pressure & 131 & 29 & 102 & 102 \\
28 Oct robustness & 22 & 0 & 22 & 18 \\
28 Oct platform & 14 & 0 & 14 & 8 \\
30 Oct platform & 146 & 96 & 50 & 44 \\
30 Oct stability & 20 & 0 & 20 & 18 \\
3 Nov speed revision & 36 & 0 & 36 & 33 \\
\bottomrule
\end{tabularx}
\end{table}

\subsection{Feedback Residuals and Acceptance Criteria}
\label{sec:residuals}

The control formulation distinguishes goal pose g, \ensuremath{\varphi} in body frame B from robot error e, \ensuremath{\psi} in goal frame G, related by \eqref{eq:6}. Statistical tables retain the signs of the logged fields: B records at P1 and P3–P5 originate from the body-frame goal pose, whereas P2 uses post-advance LiDAR distance and heading with the pre-advance visual lateral reading. We call these fields onboard feedback residuals and retain their logged signs.

Record A is acquired after coarse transfer when lateral pre-alignment is assessed; LiDAR heading pre-alignment may already have occurred. Its lateral value is projected onto the goal approach line, and its longitudinal field is empty. Record B is the final task feedback. The stage plot compares distributions of absolute lateral readings from the same tasks to describe convergence within the workflow.

The evaluated quantities are onboard feedback residuals relative to the taught state. The unified criterion is |x| \ensuremath{\leq} 10\,mm, |y| \ensuremath{\leq} 10\,mm, and |heading| \ensuremath{\leq} 1\ensuremath{^{\circ}}, with an additional strict evaluation using a 0.5\ensuremath{^{\circ}} heading limit. The former is a common-threshold retrospective evaluation of the historical records. Backend completion is counted by return code 0. Although P5 used a 20\,mm operational position gate, the primary analysis applies 10\,mm per axis at every workstation.

\subsection{Statistical and Timing Analysis}

Each executed task contributes one endpoint record. Signed feedback residuals are summarized per workstation by mean, sample standard deviation, and axis-wise RMSE. Recorded planar residual magnitude is \(\sqrt{x^2+y^2}\), from which planar RMSE, P95, and maximum are calculated; P95 uses linear interpolation. Separate limits on the two position axes define a rectangular acceptance region. Completion, first-attempt success, and residual acceptance are counted separately, with Wilson 95\% intervals \cite{r26,r28}. P4 has n = 1, so its standard deviation is undefined. Sequential tasks share the robot and learned state; the intervals describe counts in this cohort.

First-attempt success required a successful return and a logged attempts value of one; auxiliary actions within that attempt were allowed. Task duration used the recorded total, including a matched departure when available, and excluded waiting between separate upstream requests. Coarse and fine durations came from stage timestamps. For the stacked time plot, recovery/other was the total minus the recorded departure, coarse and fine durations, covering recovery, localization checks and other unassigned overhead. Four trials lacked a matched departure entry; their recorded departure contribution was zero without imputing an unobserved duration. Error bars represent the sample standard deviation of total task duration.

\section{Results and Discussion}
\label{sec:results}

\subsection{Repeated Task Completion and Residual Acceptance}

All 111 executed tasks across the two same-version sessions returned completion, with a Wilson 95\% interval of 96.7–100\%. Of these, 101 completed in one recorded attempt, corresponding to 91.0\% (84.2–95.0\%). The unified onboard residual criterion of \ensuremath{\pm}10\,mm per position axis and \ensuremath{\pm}1\ensuremath{^{\circ}} heading was met by 107 tasks, giving 96.4\% (91.1–98.6\%); 106/111 met the stricter 0.5\ensuremath{^{\circ}} heading criterion. Tables~\ref{tab:4} and~\ref{tab:5} report workstation-level signed feedback, planar residuals, and task duration.

\begin{table*}[!t]
\caption{Final recorded feedback residuals by workstation. P2 uses the composite endpoint record defined in Section~\ref{sec:residuals}.}
\label{tab:4}
\centering
\footnotesize
\setlength{\tabcolsep}{3pt}
\renewcommand{\arraystretch}{1.18}
\begin{tabularx}{\textwidth}{YYYYYY}
\toprule
Point & N & Accepted (95\% CI) & x mean \ensuremath{\pm} SD (mm) & y mean \ensuremath{\pm} SD (mm) & Heading mean \ensuremath{\pm} SD (\ensuremath{^{\circ}}) \\
\midrule
P1 & 27 & 27/27
(87.5–100.0\%) & -1.68 \ensuremath{\pm} 0.97 & -0.00 \ensuremath{\pm} 6.04 & -0.17 \ensuremath{\pm} 0.19 \\
P2 & 28 & 28/28
(87.9–100.0\%) & -0.37 \ensuremath{\pm} 0.46 & -5.21 \ensuremath{\pm} 2.15 & -0.07 \ensuremath{\pm} 0.17 \\
P3 & 28 & 28/28
(87.9–100.0\%) & -1.23 \ensuremath{\pm} 0.92 & +4.24 \ensuremath{\pm} 3.66 & -0.13 \ensuremath{\pm} 0.25 \\
P4 & 1 & 1/1
(20.7–100.0\%) & -2.20 / — & +5.89 / — & -0.06 / — \\
P5 & 27 & 23/27
(67.5–94.1\%) & -1.38 \ensuremath{\pm} 0.90 & +1.67 \ensuremath{\pm} 6.49 & -0.17 \ensuremath{\pm} 0.16 \\
All & 111 & 107/111
(91.1–98.6\%) & -1.17 \ensuremath{\pm} 0.96 & +0.21 \ensuremath{\pm} 5.95 & -0.14 \ensuremath{\pm} 0.20 \\
\bottomrule
\end{tabularx}
\end{table*}

\begin{table*}[!t]
\caption{Recorded planar residual magnitude and task duration.}
\label{tab:5}
\centering
\footnotesize
\setlength{\tabcolsep}{3pt}
\renewcommand{\arraystretch}{1.18}
\begin{tabularx}{\textwidth}{YYYYYY}
\toprule
Point & One attempt & RMSE
(mm) & P95
(mm) & Maximum (mm) & Time mean \ensuremath{\pm} SD (s) \\
\midrule
P1 & 26/27 & 6.24 & 8.60 & 10.27 & 44.4 \ensuremath{\pm} 12.8 \\
P2 & 28/28 & 5.65 & 8.61 & 9.56 & 52.3 \ensuremath{\pm} 8.4 \\
P3 & 21/28 & 5.76 & 8.47 & 9.15 & 62.2 \ensuremath{\pm} 25.7 \\
P4 & 1/1 & 6.28 & 6.28 & 6.28 & 36.4 / — \\
P5 & 25/27 & 6.79 & 11.35 & 14.86 & 46.7 \ensuremath{\pm} 13.6 \\
All & 101/111 & 6.12 & 9.97 & 14.86 & 51.4 \ensuremath{\pm} 17.6 \\
\bottomrule
\end{tabularx}
\end{table*}

All records at P1--P3 met the unified residual criterion, as did the single P4 record. P5 met it in 23/27 trials; its four exceptions were confined to lateral residuals of 10.45--14.80\,mm, with normal and heading residuals within the common limits. These four tasks completed under P5's 20\,mm operational position gate but exceeded the 10\,mm threshold used for comparison across workstations.

\subsection{Feedback Convergence with Complementary Observations}

\begin{figure*}[!t]
\centering
\includegraphics[width=\textwidth]{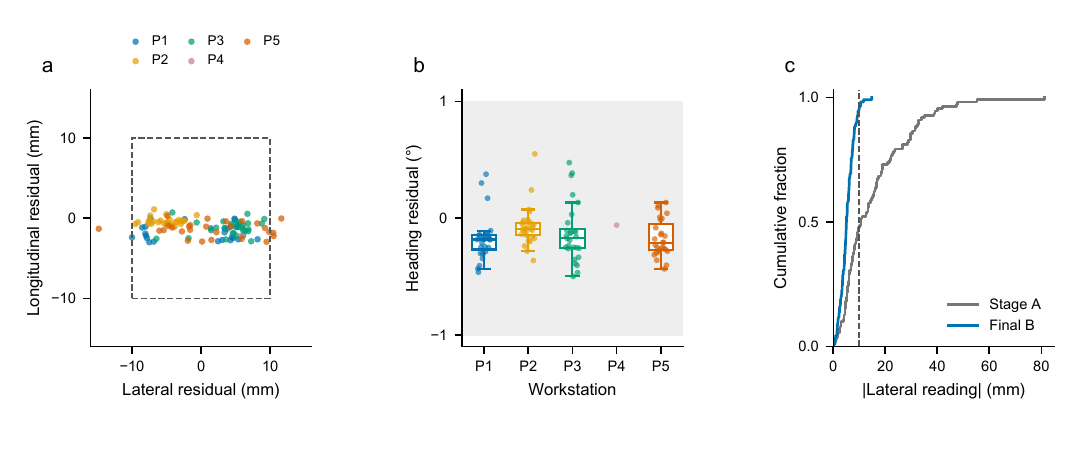}
\caption{Recorded onboard feedback residuals (n = 111). (a) Signed final position readings and the \ensuremath{\pm}10\,mm box. (b) Heading distribution and \ensuremath{\pm}1\ensuremath{^{\circ}} band; P4 has n = 1. (c) Cumulative distributions of absolute lateral readings at stages A and B for the same tasks; Section~\ref{sec:residuals} defines the fields and coordinate conventions.}
\label{fig:2}
\end{figure*}

The final recorded planar residual RMSE was 6.12\,mm, with P95 of 9.97\,mm and a maximum of 14.86\,mm. The x, y, and heading RMSE values were 1.51\,mm, 5.93\,mm, and 0.238\ensuremath{^{\circ}}, respectively. Lateral residuals accounted for 93.9\% of the planar sum of squares. We therefore regard lateral alignment as the main remaining control challenge: a differential-drive base must spend forward travel to correct this component. Fig.~\ref{fig:2} plots the endpoint residuals and the lateral readings before and after terminal control.

Across the 111 tasks, median absolute lateral readings were 10.60\,mm at stage A and 5.03\,mm in the final B records.

\subsection{Task Duration and Recovery Behavior}

\begin{figure*}[!t]
\centering
\includegraphics[width=.70\textwidth]{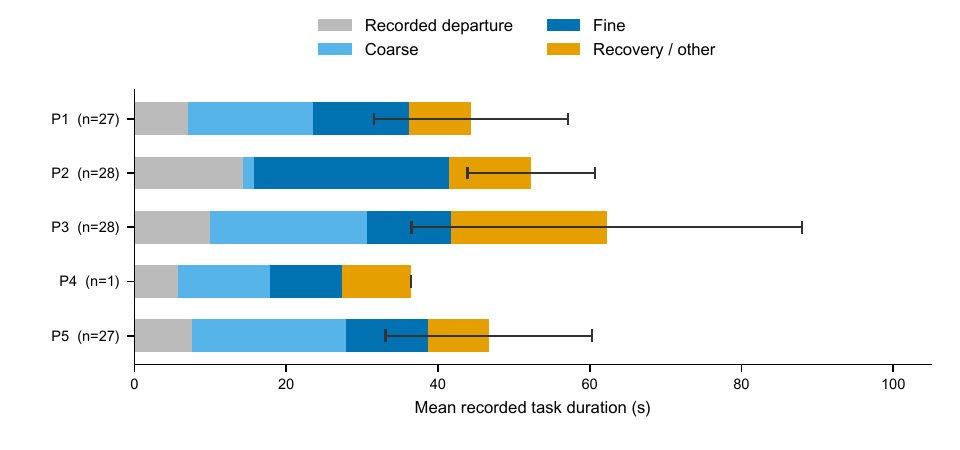}
\caption{Mean recorded duration by workstation. Segments represent recorded departure, coarse navigation, fine alignment and residual recovery/other overhead. Bars at the total-duration endpoints show one sample standard deviation. P4 has n = 1 and no estimated standard deviation. All 111 trials, including the 158.7\,s startup-localization case, are retained.}
\label{fig:3}
\end{figure*}

The recorded task duration averaged 51.4 \ensuremath{\pm} 17.6\,s, with a median of 47.7\,s. Among workstations with repeated trials, P1 had the shortest mean duration (44.4\,s), while P3 had the longest (62.2\,s) and the greatest variability (SD, 25.7\,s; Table~\ref{tab:5}). The stage decomposition in Fig.~\ref{fig:3} places these differences in the full operating cycle, including departure, coarse navigation, fine alignment and recovery or other overhead. Because the routes and approach conditions differed across workstations, these totals characterize operational task time rather than a controlled comparison of servo speed.

All ten tasks with two attempts ultimately completed: seven at P3, two at P5, and one at P1. At P3, lateral gate exceedance triggered all seven retries, with first-endpoint absolute residuals of 12.38--68.19\,mm; longitudinal and heading readings remained within the common limits. Six lateral readings were negative and one was positive. Six recovery records included lateral pre-alignment; one did not trigger it after retreat. The longest task, 158.7\,s, also included startup tag-based relocalization. Repeated approaches and associated overhead increased P3 cycle time, whereas P5 final lateral exceedances remained within its relaxed operational gate.

\begin{figure*}[!t]
\centering
\includegraphics[width=\textwidth]{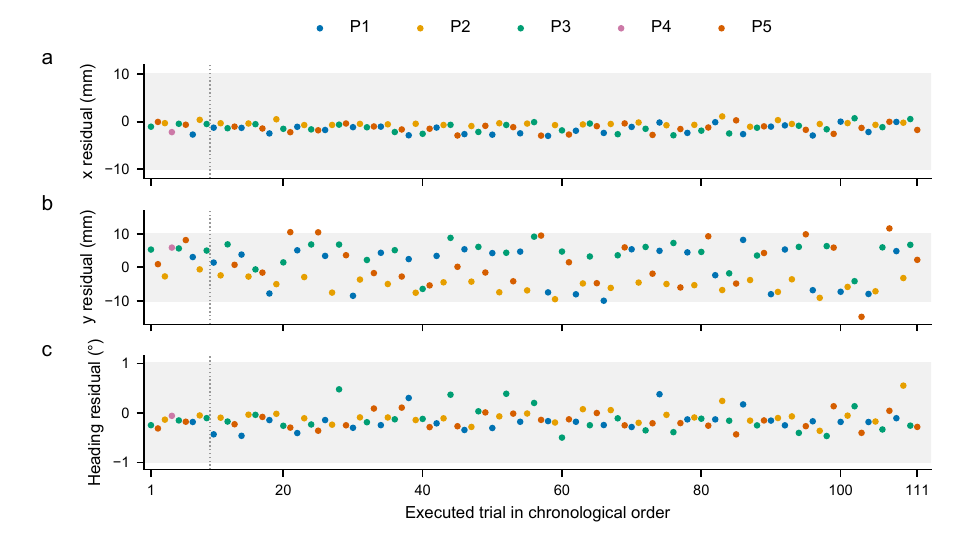}
\caption{Final recorded feedback residuals across two sessions on the same day: (a) longitudinal, (b) lateral, and (c) heading. The dotted boundary separates nine startup-session tasks from 102 pressure-test tasks. The horizontal axis is task order, and shaded bands indicate the common thresholds.}
\label{fig:4}
\end{figure*}

\subsection{Workstation Biases and Their Interpretation}
The longitudinal feedback mean was $-1.17$\,mm, with a negative mean at all five workstations. The sign has different meanings across records: P1 and P3--P5 store body-frame goal readings, whereas P2 stores a surface-distance difference. It must therefore be interpreted using the corresponding frame and measurement definition before drawing conclusions about physical overshoot.

Mean lateral residuals of $-5.21$\,mm at P2 and $+4.24$\,mm at P3 partly cancel in the pooled mean of $+0.21$\,mm. The near-zero pooled value masks workstation-specific offsets. We retain P2's composite endpoint record to report the feedback available during operation. Its lateral value precedes the final advance, so we interpret it as workflow feedback rather than a simultaneous measurement of final pose. P3 records include updates to the lateral-action model. Predictive stopping, final pulses, turning motion, the taught reference and calibration can affect endpoint readings; their contributions remain unresolved in these aggregate records. Fig.~\ref{fig:4} shows variation across the two same-day sessions in task order.

\subsection{Limitations}
Independent physical endpoint accuracy was not measured; the reported residuals compare onboard feedback with taught observations. The trials cover one site and two sessions on one day, with a single P4 record. Sequential tasks share the robot, and learned runtime state continued to update. P2's composite endpoint lacks a visual lateral measurement after the final advance, while stage A/B readings differ in frame and timing. A simultaneous external pose measurement at the final stop would resolve whether the recorded residuals track physical alignment, particularly at P2.

RPP collision checking and active collision\_monitor polygons were disabled under emergency-stop supervision, so completion rate does not quantify unattended operating safety. The evaluation concerns the integrated workflow; attributing gains to individual modules would require a Nav2-only comparison or module ablations.

\section{Conclusion}
A fitted cabinet line leaves translation along the surface unobserved. We resolve that component with the tag-center bearing and use the resulting local feedback to guide lateral pre-alignment and continuous visual servoing. All 111 tasks using the same navigation and docking code completed, including 101 on the first attempt. Onboard longitudinal, lateral and heading residual RMSE values were 1.51\,mm, 5.93\,mm and $0.238^\circ$, with 96.4\% joint residual acceptance within $\pm10$\,mm per axis and $\pm1^\circ$ heading. An ordinary cabinet front and a fixed tag thus form the local reference for repeated docking at the taught workstation pose.

\appendices
\section{Implementation Parameters}
\label{app:params}
Table~\ref{tab:parameters} summarizes key operating settings after the 3 November revisions. Section~\ref{sec:versions} gives the settings used for the 2 November primary cohort.
\begin{table}[!t]
\caption{Key operating settings}
\label{tab:parameters}\centering\footnotesize
\begin{tabularx}{\columnwidth}{lY}\toprule
Item & Value \\ \midrule
Image offset / matching & $0.055$ / $0.30$\,s \\
Staging setback & $b=0.30$\,m \\
Control rate & 10\,Hz \\
Heading floor / slope & $2^\circ$; $100\pi/180$\,rad/m \\
Predictive stopping & $\ell_{\mathrm{stop}}=4$\,mm; $\tau_{\mathrm{stop}}=0.35$\,s \\
Observation age / outage & $0.6$ / $3$\,s \\
Approach timeout / travel & 40\,s / $2.0$\,m \\
Operational position gate & P1--P4: 10\,mm; P5: 20\,mm \\
\bottomrule\end{tabularx}\end{table}

\subsection{Localization, Handover, and Pre-Alignment}
Global localization uses the occupancy grid and pose graph stored by SLAM Toolbox \cite{r2}. A stationary LiDAR scan is aligned to the map through distance-field search and local pose refinement, followed by localization verification. Historical poses, manual hints, or tag-derived hints assist candidate selection in geometrically ambiguous scenes. Before the first docking task after each system boot, the controller checks localization against an observed workstation tag. It searches by rotation when necessary and initializes localization again if the discrepancy exceeds 0.25 m or 10\ensuremath{^{\circ}}. Failure to obtain a valid tag-based check prevents departure.

Path tracking uses the Regulated Pure Pursuit controller \cite{r5}. Let $({x}_{j},{y}_{j},{\theta }_{j})$ denote the taught docking goal for workstation $j$ in the map frame. The staging pose is defined as

\begin{equation}
\begin{aligned}\begin{bmatrix}x_s\\y_s\end{bmatrix}&=\begin{bmatrix}x_j\\y_j\end{bmatrix}-b\begin{bmatrix}\cos\theta_j\\\sin\theta_j\end{bmatrix}+o_j\begin{bmatrix}-\sin\theta_j\\\cos\theta_j\end{bmatrix},\\ \theta_s&=\theta_j.\end{aligned}
\label{eq:1}
\end{equation}

Here, b = 0.30 m is the backward offset along the goal heading. The lateral offset ${o}_{j}$ is learned separately for each workstation from its observed arrival offset. Successive estimates receive equal weights of 0.5, and the stored offset is limited to \ensuremath{\pm}0.15 m.

Early handover requires a remaining Nav2 path distance below 0.15 m after a previous feedback value has exceeded 0.20 m, excluding an initial zero-distance report. The selected tag must have been received within 0.4\,s, goal-frame lateral error must be within 50\,mm, heading within 6\ensuremath{^{\circ}}, odometry angular speed within 0.1\,rad/s, and remaining forward distance within 0.12–0.40 m.

Cancellation must finish before local motion begins. The controller then requires odometry newer than 0.3\,s, speeds below 0.01\,m/s and 0.02\,rad/s, and no navigation velocity command within the preceding 0.5\,s. These conditions must persist for 0.3\,s. If early visual handover is unavailable, a remaining-path threshold of 0.06 m permits a separately controlled final heading turn, subject to clearance checks.

Before continuous approach, LiDAR heading alignment precedes evaluation of lateral error in the goal frame, avoiding heading-induced contamination of body-frame lateral readings. Lateral pre-alignment is triggered above 15\,mm at workstations 1 and 3, and above 30\,mm at workstations 2, 4, and 5. A turn–translate–realign maneuver moves the robot toward the approach line when the observed turning region and translation corridor permit.

Pre-alignment compensates for turning-induced displacement with a workstation-specific affine model relating commanded lateral shift to measured error removal. The command subtracts the fitted offset from the required correction and divides by the fitted gain, with magnitude limited to 0.25 m. A least-squares fit uses up to 20 recent maneuvers and becomes active after four samples. The gain is bounded to 0.5–1.3 and the offset to \ensuremath{\pm}80\,mm. When command span is below 30\,mm, gain is fixed at 0.85 and only the offset is fitted. Before enough samples are available, a bounded multiplicative gain is used.

Configured adjacent-workstation transfers reduce repeated global planning. The 1\ensuremath{\rightarrow}2 transfer uses a reverse arc of radius 0.225 m after retreating to its start position. The 2\ensuremath{\rightarrow}3 transfer retreats approximately 0.52 m and turns approximately 88\ensuremath{^{\circ}}; the 3\ensuremath{\rightarrow}4 transfer translates along the shared surface after reorientation. Transfer distances derive from workstation geometry and stored corrections. Each transfer is followed by local observation, pre-alignment when necessary, and the same terminal servo controller.

\subsection{Observation Gating and Teaching}
Relative pose estimation uses the cabinet geometry seen by LiDAR and the tag bearing seen by the camera. The sensors are asynchronous. We assign each image an effective acquisition time ${t}_{c}={t}_{\mathrm{img}}-\Delta t$, with $\Delta t=0.055\,\mathrm{s}$, and select the nearest LiDAR scan within $0.30\,\mathrm{s}$. Each observation is transformed into odometry frame $O$ at its own acquisition time. This aligns frames using the fixed image offset and robot motion; individual LiDAR beams are not deskewed.

LiDAR points within 0.60 m of the initial tag position form the candidate surface region. RANSAC \cite{r18} uses 200 candidate point pairs and a 10\,mm inlier threshold, followed by three orthogonal least-squares refinement iterations. Let I denote the retained inlier set and ${p}_{i}$ a LiDAR point. The fitted line has centroid c and unit normal n, with the objective in \eqref{eq:2}.

The normal points toward the LiDAR. Refinement uses a 20\,mm residual threshold and a window extending 0.35 m on either side of the projected tag position. A valid fit requires at least 60 inliers, a segment length of at least 0.55 m, and a root-mean-square residual no greater than 8\,mm.

OpenCV detects tag36h11 AprilTags \cite{r8} in infrared images after intensity normalization using the 1st and 99.5th percentiles. AprilTag corner refinement follows detection, with a local search around the previous tag position and full-image search if needed. IPPE \cite{r10} provides the initial tag position and reprojection residual. The Robot 2 configuration accepts estimated tag ranges of 0.12–1.50 m and reprojection errors no greater than 3.5 pixels. Tangential position is then determined by ray–plane intersection.

To teach a target, we collect surface observations with the robot at the desired pose. A teaching request collects 200 usable frames by default. Translation is averaged arithmetically and orientation circularly, yielding ${}^{B^*}\bm T_F$, where B* is the taught body frame. We also store mean normal distance, tangential position, tag identity and observation variability. In ray mode, standard deviation must stay within 1\,mm for normal distance and each tag-position component, and within 0.05\ensuremath{^{\circ}} for surface orientation. Held-tag frames and ambiguous detections are excluded. Tag-orientation variability is recorded but does not gate ray-mode teaching.

Goal estimates are rejected when consecutive odometry-frame targets differ by more than 20\,mm or 1\ensuremath{^{\circ}}. During brief tag loss, perception can retain the last valid tag position in odometry for up to 1\,s while updating the surface from LiDAR. Compensation for a retained tag uses the surface distance at its last actual detection. These frames are marked as held-tag observations and are excluded from teaching, continuous visual control, and terminal visual acceptance.

\subsection{Continuous Servoing and Endpoint Sampling}
The initial lateral offset selects the steering schedule. Below 40\,mm, the lateral gain is 12 rad/m, the heading cap is 10\ensuremath{^{\circ}}, and the far-field angular-velocity limit is 0.06\,rad/s. Otherwise, these values are 7 rad/m, 18\ensuremath{^{\circ}}, and 0.10\,rad/s. Within 0.10 m, the angular-velocity limit becomes 0.04\,rad/s. The heading gain is 1.2\,s\ensuremath{^{-1}}, the distance gain is 0.4\,s\ensuremath{^{-1}}, and the minimum forward speed is 0.012\,m/s. Maximum speed is 0.05\,m/s at workstations 1–4 and 0.06\,m/s at workstation 5. During the last 80\,mm, the desired heading becomes zero, so lateral steering gives way to heading alignment.

The lateral steering target ${\eta }_{j}$ compensates for repeatable reading drift over the final straight segment. It is enabled only at workstation 2. After an accepted visual approach, the measured lateral drift from straight-segment entry to the taught endpoint updates the opposite steering offset with weight 0.5. The update is skipped if entry heading exceeds 3\ensuremath{^{\circ}} or drift exceeds 15\,mm, and the offset is limited to \ensuremath{\pm}12\,mm. This correction is separate from the distance-dependent observation compensation in \eqref{eq:4}.

Here, the preceding linear-velocity command determines the predicted stopping distance, with ${\ell }_{\mathrm{stop}}$ = 4\,mm and ${\tau }_{\mathrm{stop}}$ = 0.35\,s. After stopping, up to four adjustment rounds target 3\,mm longitudinal error and 1\ensuremath{^{\circ}} heading error. Each round first requests five valid observations. Heading error is corrected before longitudinal error; longitudinal correction uses 0.01\,m/s pulses lasting 0.15–1.5\,s. Pulse gain adapts to LiDAR-measured displacement and is bounded between 0.3 and 2.0. The pulse loop targets a remaining displacement within 1\,mm. At workstation 2, longitudinal adjustment at the taught pose is skipped when its error is already within 10\,mm, because the subsequent advance sets the final distance.

The continuous controller uses actual-tag observations received within 0.6\,s. Without such observations, it commands zero velocity and waits; an outage exceeding 3\,s ends the approach attempt. Each attempt also has a 40\,s servo timeout and a 2.0 m cumulative travel limit. The deployed task controller allows one servo pass per invocation. If the endpoint fails acceptance and a retry is permitted, the robot retreats to the staging distance, performs lateral pre-alignment with an 8\,mm threshold, and makes one further pass. Missing target observations after navigation can also trigger tag-based relocalization before another approach.

Terminal visual readings average the longitudinal, lateral and heading components of valid observations from the selected workstation. Held-tag frames are excluded. A settled five-frame reading from the final adjustment is used directly when it meets the adjustment targets. Otherwise, the controller requests eight observations within 4\,s. If fewer arrive, the nonempty partial sample is retained and its actual count is recorded.

At workstation 2, acceptance at the taught pose is followed by LiDAR-guided advance toward a target surface distance of 0.3772 m. The robot advances at 0.04\,m/s toward a point 12\,mm short of that distance, with predictive braking, then applies adaptive pulses targeting 1\,mm residual. After this motion, ten LiDAR scans are requested; componentwise medians of valid surface fits remeasure distance and heading. The final longitudinal residual is measured surface distance minus 0.3772 m; the heading residual is measured surface orientation minus its taught value. Because the tag is no longer visible at this close endpoint, the lateral component retains its pre-advance visual reading. These final components are checked again against the workstation thresholds, while the visual-pose record is retained separately.

\subsection{State Consistency and Startup Checks}
The docking backend connects to upstream scheduling through standard DockRobot and UndockRobot actions. At task start, it checks the stored state again: odometry displacement beyond 50\,mm, rotation beyond 5\ensuremath{^{\circ}}, or a map-position discrepancy exceeding 0.35 m from the recorded workstation invalidates that state and triggers planning from the current pose. Undocking motion requires confirmation of the corresponding workstation. A stop signal commands zero velocity before terminating the active action, providing a definite terminal state to the scheduler.

Startup checks include both localization availability and data-link availability. A link-check service attempts Ethernet renegotiation when the link rate falls below gigabit speed, followed by verification of infrared-image reception; this recovery action addresses link-negotiation faults. The system checks gigabit networking, infrared-image reception, and intercomputer clock offset, with the boot script waiting for an ECU–Thor difference below 50 ms; a timeout is logged as a warning and startup continues. The first task after boot then verifies map localization against tag identity and pose. This check identifies geometric matching ambiguities in symmetric scenes and returns a failure state when trustworthy observations are unavailable.

\FloatBarrier


\begin{thebibliography}{28}

\bibitem{r1} C. Cadena et al., ``Past, Present, and Future of Simultaneous Localization and Mapping: Toward the Robust-Perception Age,'' \emph{IEEE Trans. Robot.}, vol. 32, no. 6, pp. 1309--1332, 2016. \url{https://doi.org/10.1109/tro.2016.2624754}

\bibitem{r2} S. Macenski, and I. Jambrečić, ``SLAM Toolbox: SLAM for the dynamic world,'' \emph{J. Open Source Softw.}, vol. 6, no. 61, Art. no. 2783, 2021. \url{https://doi.org/10.21105/joss.02783}

\bibitem{r3} S. Macenski, F. Martín, R. White, and J. G. Clavero, ``The Marathon 2: A Navigation System,'' in \emph{Proc. IEEE/RSJ Int. Conf. Intell. Robots Syst. (IROS)}, 2020, pp. 2718--2725. \url{https://doi.org/10.1109/iros45743.2020.9341207}

\bibitem{r4} G. Williams, P. Drews, B. Goldfain, J. M. Rehg, and E. A. Theodorou, ``Information-Theoretic Model Predictive Control: Theory and Applications to Autonomous Driving,'' \emph{IEEE Trans. Robot.}, vol. 34, no. 6, pp. 1603--1622, 2018. \url{https://doi.org/10.1109/tro.2018.2865891}

\bibitem{r5} S. Macenski, S. Singh, F. Martín, and J. Ginés, ``Regulated pure pursuit for robot path tracking,'' \emph{Auton. Robots}, vol. 47, no. 6, pp. 685--694, 2023. \url{https://doi.org/10.1007/s10514-023-10097-6}

\bibitem{r6} S. Ç. Tekkök, B. Bostanci, M. E. Söyünmez, and P. Oğuz Ekim, ``A Novel Docking Algorithm Based On The LiDAR And The V-shape Features,'' \emph{Eur. J. Sci. Technol.}, no. 26, pp. 35--40, 2021. \url{https://doi.org/10.31590/ejosat.947521}

\bibitem{r7} Y. Liu, ``A Laser Intensity Based Autonomous Docking Approach for Mobile Robot Recharging in Unstructured Environments,'' \emph{IEEE Access}, vol. 10, pp. 71165--71176, 2022. \url{https://doi.org/10.1109/access.2022.3188099}

\bibitem{r8} E. Olson, ``AprilTag: A robust and flexible visual fiducial system,'' in \emph{Proc. IEEE Int. Conf. Robot. Autom. (ICRA)}, 2011, pp. 3400--3407. \url{https://doi.org/10.1109/icra.2011.5979561}

\bibitem{r9} J. Wang, and E. Olson, ``AprilTag 2: Efficient and robust fiducial detection,'' in \emph{Proc. IEEE/RSJ Int. Conf. Intell. Robots Syst. (IROS)}, 2016, pp. 4193--4198. \url{https://doi.org/10.1109/iros.2016.7759617}

\bibitem{r10} T. Collins, and A. Bartoli, ``Infinitesimal Plane-Based Pose Estimation,'' \emph{Int. J. Comput. Vis.}, vol. 109, no. 3, pp. 252--286, 2014. \url{https://doi.org/10.1007/s11263-014-0725-5}

\bibitem{r11} J. H. Richter, D. Bohlig, A. Nüchter, and K. Schilling, ``Advanced Edge Detection of AprilTags for Precise Docking Maneuvers of Mobile Robots,'' \emph{IFAC-PapersOnLine}, vol. 55, no. 8, pp. 117--123, 2022. \url{https://doi.org/10.1016/j.ifacol.2022.08.020}

\bibitem{r12} R. C. Luo, C. T. Liao, and S. C. Lin, ``Multi-sensor fusion for reduced uncertainty in autonomous mobile robot docking and recharging,'' in \emph{Proc. IEEE/RSJ Int. Conf. Intell. Robots Syst. (IROS)}, 2009, pp. 2203--2208. \url{https://doi.org/10.1109/iros.2009.5354445}

\bibitem{r13} F. Jia, M. Afaq, B. Ripka, Q. Huda, and R. Ahmad, ``Vision- and Lidar-Based Autonomous Docking and Recharging of a Mobile Robot for Machine Tending in Autonomous Manufacturing Environments,'' \emph{Appl. Sci.}, vol. 13, no. 19, Art. no. 10675, 2023. \url{https://doi.org/10.3390/app131910675}

\bibitem{r14} Y. Dai, and K. Lee, ``Multi-Sensor Fusion for Autonomous Mobile Robot Docking: Integrating LiDAR, YOLO-Based AprilTag Detection, and Depth-Aided Localization,'' \emph{Electronics}, vol. 14, no. 14, Art. no. 2769, 2025. \url{https://doi.org/10.3390/electronics14142769}

\bibitem{r15} A. Yilmaz, and H. Temeltas, ``A multi-stage localization framework for accurate and precise docking of autonomous mobile robots (AMRs),'' \emph{Robotica}, vol. 42, no. 6, pp. 1885--1908, 2024. \url{https://doi.org/10.1017/s0263574724000602}

\bibitem{r16} F. Chaumette, and S. Hutchinson, ``Visual servo control. I. Basic approaches,'' \emph{IEEE Robot. Autom. Mag.}, vol. 13, no. 4, pp. 82--90, 2006. \url{https://doi.org/10.1109/mra.2006.250573}

\bibitem{r17} J. J. Park, and B. Kuipers, ``A smooth control law for graceful motion of differential wheeled mobile robots in 2D environment,'' in \emph{Proc. IEEE Int. Conf. Robot. Autom. (ICRA)}, 2011, pp. 4896--4902. \url{https://doi.org/10.1109/icra.2011.5980167}

\bibitem{r18} M. A. Fischler, and R. C. Bolles, ``Random sample consensus: a paradigm for model fitting with applications to image analysis and automated cartography,'' \emph{Commun. ACM}, vol. 24, no. 6, pp. 381--395, 1981. \url{https://doi.org/10.1145/358669.358692}

\bibitem{r19} M. Kutan, and H. Çiğ, ``A Multi-Stage 2D LiDAR-Based Reactive Docking Framework for Precision Alignment of Industrial Autonomous Mobile Robots,'' \emph{Dicle Üniversitesi Mühendislik Fakültesi Mühendislik Dergisi}, vol. 17, no. 3, 2026. \url{https://doi.org/10.24012/dumf.1938367}

\bibitem{r20} M. Krogius, A. Haggenmiller, and E. Olson, ``Flexible Layouts for Fiducial Tags,'' in \emph{Proc. IEEE/RSJ Int. Conf. Intell. Robots Syst. (IROS)}, 2019, pp. 1898--1903. \url{https://doi.org/10.1109/iros40897.2019.8967787}

\bibitem{r21} Y. H. Huang, and C. T. Lin, ``Indoor Localization Method for a Mobile Robot Using LiDAR and a Dual AprilTag,'' \emph{Electronics}, vol. 12, no. 4, Art. no. 1023, 2023. \url{https://doi.org/10.3390/electronics12041023}

\bibitem{r22} P. Furgale, and T. D. Barfoot, ``Visual teach and repeat for long-range rover autonomy,'' \emph{J. Field Robot.}, vol. 27, no. 5, pp. 534--560, 2010. \url{https://doi.org/10.1002/rob.20342}

\bibitem{r23} F. Chaumette, and S. Hutchinson, ``Visual servo control. II. Advanced approaches,'' \emph{IEEE Robot. Autom. Mag.}, vol. 14, no. 1, pp. 109--118, 2007. \url{https://doi.org/10.1109/mra.2007.339609}

\bibitem{r24} Z. Zhang, ``A flexible new technique for camera calibration,'' \emph{IEEE Trans. Pattern Anal. Mach. Intell.}, vol. 22, no. 11, pp. 1330--1334, 2000. \url{https://doi.org/10.1109/34.888718}

\bibitem{r25} S. Garrido-Jurado, R. Muñoz-Salinas, F. J. Madrid-Cuevas, and M. J. Marín-Jiménez, ``Automatic generation and detection of highly reliable fiducial markers under occlusion,'' \emph{Pattern Recognit.}, vol. 47, no. 6, pp. 2280--2292, 2014. \url{https://doi.org/10.1016/j.patcog.2014.01.005}

\bibitem{r26} L. D. Brown, T. T. Cai, and A. DasGupta, ``Interval Estimation for a Binomial Proportion,'' \emph{Statist. Sci.}, vol. 16, no. 2, pp. 101--133, 2001. \url{https://doi.org/10.1214/ss/1009213286}

\bibitem{r27} J. Borenstein, and L. Feng, ``Measurement and correction of systematic odometry errors in mobile robots,'' \emph{IEEE Trans. Robot. Autom.}, vol. 12, no. 6, pp. 869--880, 1996. \url{https://doi.org/10.1109/70.544770}

\bibitem{r28} E. B. Wilson, ``Probable Inference, the Law of Succession, and Statistical Inference,'' \emph{J. Amer. Statist. Assoc.}, vol. 22, no. 158, pp. 209--212, 1927. \url{https://doi.org/10.1080/01621459.1927.10502953}

\end{thebibliography}
\end{document}